\pdfoutput=1
\documentclass[11pt]{article}

\usepackage[preprint]{acl}

\usepackage{times}
\usepackage{latexsym}

\usepackage[T1]{fontenc}
\usepackage[utf8]{inputenc}

\usepackage{microtype}

\usepackage{inconsolata}

\usepackage{graphicx}
\usepackage{multirow}
\usepackage{array}
\usepackage{booktabs}
\usepackage{makecell}
\title{Opportunities and Challenges of LLMs in Education: An NLP Perspective}

 \author{Sowmya Vajjala$^1$, Bashar Alhafni$^2$, Stefano Bannò$^3$, \\ {\bf Kaushal Kumar Maurya$^2$, Ekaterina Kochmar$^2$} \\
  $^1$National Research Council, Canada, $^2$MBZUAI, $^3$University of Cambridge\\
  \texttt{sowmya.vajjala@nrc-cnrc.gc.ca, bashar.alhafni@mbzuai.ac.ae,}  \\ \texttt{sb2549@eng.cam.ac.uk,\{kaushal.maurya, ekaterina.kochmar\}@mbzuai.ac.ae} \\
}

\date{}

\begin{document}
\maketitle
\begin{abstract}
Interest in the role of large language models (LLMs) in education is increasing, considering the new opportunities they offer for teaching, learning, and assessment. In this paper, we examine the impact of LLMs on educational NLP in the context of two main application scenarios: {\em assistance} and {\em assessment}, grounding them along the four dimensions -- reading, writing, speaking, and tutoring. We then present the new directions enabled by LLMs, and the key challenges to address. We envision that this holistic overview would be useful for NLP researchers and practitioners interested in exploring the role of LLMs in developing language-focused and NLP-enabled educational applications of the future. 
\end{abstract}

\section{Introduction}
\label{sec:intro}
Large language models (LLMs) have demonstrated remarkable capabilities across various tasks within and beyond NLP. The rapid adoption of LLMs and generative AI by EdTech companies such as Duolingo~\cite{naismith2023automated} and Grammarly \cite{raheja-etal-2023-coedit,raheja-etal-2024-medit} and the development of fine-tuned models for educational use cases such as LearnLM \cite{team2024learnlm} are some examples of real-world impact in Education domain. The NLP community has a long history in this area, especially on problems such as automated essay scoring, grammatical error correction, and text simplification, to name a few. Naturally, there is a huge interest in using LLMs for educational applications within the community. While LLMs have undoubtedly caused a paradigm shift in this area, enabling new opportunities in writing assistance, personalization, and interactive teaching and learning, among other tasks, they also present novel challenges. %, particularly regarding the ethical considerations in integrating them into educational settings, fair assessment, and evaluation. 
In this paper, we delve into the opportunities and challenges presented by LLMs for educational applications by considering the use cases involving language, and instruction in natural language, and connect the recent developments to past NLP research in this area, outlining the path ahead. 

\begin{figure}[t]
    \centering
\includegraphics[width=\linewidth]{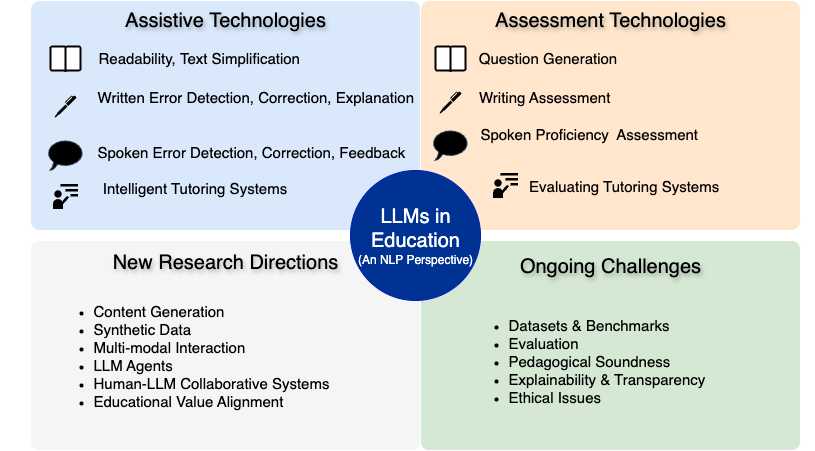}
    \caption{Overview of the paper.}
    %We cover assistive and assessment technologies in terms of reading, writing, speaking, and tutoring.}
%The paper is organized into four key components, providing a comprehensive summary of existing research while highlighting newly emerging directions and ongoing challenges.}
    \label{fig:overview}
\end{figure}

We group the state of the art into two main topics: {\em assistive technologies} -- meant to support students and teachers (\S\ref{sec:assistive}) -- and {\em assessment technologies} -- meant to assess the performance of students (\S\ref{sec:assessment}). Within each, we discuss the role of NLP and LLMs across specific aspects of education -- reading, writing, speaking, and general tutoring. We then turn to some of the new directions enabled by LLMs in NLP in this area (\S\ref{sec:newdirections}), point to some ongoing challenges (\S\ref{sec:challenges}), %, bringing in insights from other related disciplines such as applied linguistics and educational psychology, where relevant. and
and summarize our key insights (\S\ref{sec:summary}). 
%Despite the two decade-long history of NLP research and practice in this area, we do not yet have a comprehensive overview covering the different applications of NLP in education. While we will not survey the entire space in this paper, we aim to give a brief historical context and cover the recent developments in the field,  outlining some of the emerging trends in this direction and putting NLP research in the perspective of the broader influence of LLMs in Education. 

While there are other related surveys on this topic in other disciplines \cite[e.g.,][]{yan2024practical}, we focus specifically on the development of language-based educational applications, and how other existing NLP methods fit into the current scenario (See Appendix~\ref{sec:relatedwork} for a discussion on related recent surveys outside NLP). In terms of scope, as we focus on language use, we exclude topics such as learning analytics, development of student models, measuring long-term educational outcomes, interactive classroom technologies, user studies, and similar. 

\section{Assistive Technologies}
\label{sec:assistive}
We refer to the NLP problems focused on supporting learners and/or instructors as {\em assistive technologies}, and discuss them by splitting them into four groups: writing, speaking, reading, and general tutoring. Note that we focus on the recent developments and refer to the relevant surveys for the pre-LLM research, where needed. 

\subsection{Writing}
\label{sec:written-gec}
% GEC and GED, error explanation / feedback generation.
Assistive technologies for writing primarily focus on \textbf{Grammatical Error Detection (GED) and Correction (GEC)}, 
% Grammatical error correction (GEC) aims to correct errors in text, including grammatical, orthographic, and semantic errors, among other types of linguistic errors.
% A closely related task is grammatical error detection (GED), which focuses on identifying and classifying such errors.
both of which have a long-standing pedagogical value in writing assistance tool development. 
%designed to support both native (L1) and second-language (L2) learners
GEC has witnessed significant progress over the past two decades through the organization of several shared tasks \cite[\textit{inter alia}]{ng-etal-2014-conll,bryant-etal-2019-bea,masciolini-etal-2025-multigec}. For a comprehensive overview of the GEC literature, see the survey by \newcite{bryant-etal-2023-grammatical}. While GEC has received much of the attention, GED has also evolved as a stand-alone task \cite[\textit{inter alia}]{tetreault-chodorow-2008-ups,leacock-etal-2014-automated,rei-yannakoudakis-2016-compositional}. 
% Early approaches to GEC relied on rule-based systems and statistical models \cite[\textit{interalia}]{chodorow-etal-2007-detection,kochmar-etal-2012-hoo,felice-etal-2014-grammatical,junczys-dowmunt-grundkiewicz-2014-amu}. With the rise of neural architectures, the field shifted toward sequence-to-sequence models \cite[\textit{interalia}]{junczys-dowmunt-etal-2018-approaching,yuan-etal-2019-neural,kaneko-etal-2020-encoder,zhou-etal-2023-improving-seq2seq} and text-editing approaches \cite[\textit{interalia}]{malmi-etal-2019-encode,stahlberg-kumar-2020-seq2edits,omelianchuk-etal-2020-gector,alhafni2025enhancing}.

% While GEC has received much of the attention, GED has also evolved as a stand-alone task \cite{leacock-etal-2014-automated}. Early approaches focused on detecting specific error types \cite[\textit{interalia}]{han-etal-2004-detecting,tetreault-chodorow-2008-ups,lee-seneff-2008-correcting,cahill-etal-2013-detecting,kochmar-briscoe-2014-detecting}, while neural approaches have primarily targeted binary GED without being limited to any particular type of error \cite[\textit{interalia}]{rei-yannakoudakis-2016-compositional,rei-yannakoudakis-2017-auxiliary,bell-etal-2019-context,kaneko2019}. Additionally, GED has also been used an auxiliary task to improve GEC \cite{yuan-etal-2019-neural,zhao-etal-2019-improving,yuan-etal-2021-multi,alhafni-etal-2023-advancements} and to support fine-grained error analysis in system evaluation \cite{bryant-etal-2017-automatic,choshen-etal-2020-classifying,choshen2021,belkebir-habash-2021-automatic}.

Several recent studies have applied LLMs to (mainly English) GEC, comparing prompting methods along two dimensions: strategy (e.g., zero-shot, few-shot, chain-of-thought) and design (e.g., fluency-oriented vs. minimal edits). So far, few-shot prompting tends to outperform zero-shot, while chain-of-thought shows no clear benefit \cite{fang2023chatgpt,coyne2023analyzingperformancegpt35gpt4,wu2023chatgptgrammarlyevaluatingchatgpt,loem-etal-2023-exploring,davis-etal-2024-prompting,katinskaia-yangarber-2024-gpt,omelianchuk-etal-2024-pillars}. In terms of performance, LLMs often outperform state-of-the-art models on some benchmarks such as JFLEG \cite{napoles-etal-2017-jfleg} due to their strength in generating fluent rewrites, but underperform on larger benchmarks like CoNLL-2014 \cite{ng-etal-2014-conll} and BEA-2019 \cite{bryant-etal-2019-bea}, which prioritize precision and minimal edits. This reflects the difficulty of controlling LLMs to make minimal, targeted corrections, which is essential in educational applications where the goal is to guide learners in revising their own errors while preserving intent \cite{nicholls:2003}.

LLMs have also been leveraged for \textbf{Grammatical Error Explanation (GEE)}, a task that combines GED and GEC to generate natural language explanations of learner errors. Recent work has introduced methods to guide LLMs in producing such explanations using detected edits %, along with datasets like XGEC and GMEG-EXP, which extend samples of existing GEC corpora with human- and LLM-generated explanations 
\cite{kaneko-okazaki-2024-controlled,lopez-cortez-etal-2024-gmeg}. \newcite{song-etal-2024-gee} evaluated LLMs on GEE in English, German, and Chinese, showing that models often struggle to identify and explain errors, though performance improves when edits are included in the prompt. There is a growing interest in GEE for other languages as well \cite{ye2025excgec,maity2025leveraging}. LLMs have also been shown to be useful in providing feedback on other aspects of language assessment such as vocabulary usage ~\cite{ortiz-zambrano-etal-2024-enhancing, banno2025exploitingenglishvocabularyprofile}, discourse coherence ~\cite{naismith-etal-2023-automated} and analytical assessment of written texts ~\cite{banno2024gpt, stahl-etal-2024-exploring}, indicating the growing interest in this direction. Note that some previous work on feedback comment generation for writing also pursued similar goals but preceded the widespread adoption of LLMs \cite{nagata-2019-toward,nagata-etal-2020-creating,hanawa-etal-2021-exploring,nagata-etal-2021-shared}. 
%This is now a topic of increasing interest with new research emerging for other languages as well \cite{ye2025excgec,maity2025leveraging}

% GEE: 
% 1) \cite{kaneko-okazaki-2024-controlled}: https://aclanthology.org/2024.lrec-main.350/
% 2) \cite{lopez-cortez-etal-2024-gmeg}: https://aclanthology.org/2024.lrec-main.688/
% 3) \cite{fei-etal-2023-enhancing}: https://aclanthology.org/2023.acl-long.413/
% 4) \cite{song-etal-2024-gee}: https://aclanthology.org/2024.findings-naacl.49/
% 5) https://arxiv.org/pdf/2502.15261

\subsection{Speaking}
As with writing, a common application for supporting learners in speech is \textbf{spoken GEC}. However, compared to written GEC which typically works with well-formed inputs where punctuation and capitalization can aid in error detection, 
% GEC in written language is a well-established research domain, supported by shared tasks such as CoNLL-2014 \cite{ng2014shared}, BEA-2019 \cite{bryant2019bea}, and MULTIGEC-2025 \cite{multigec2024}. These initiatives have resulted in robust benchmark datasets and high-performing models capable of correcting a wide spectrum of grammatical errors in text. However, written GEC typically operates under relatively clean and structured conditions: inputs are well-formed, sentences are complete, and cues such as punctuation and capitalization aid in error detection.
spoken GEC presents a distinct set of challenges which significantly complicate the task of identifying and correcting grammatical errors in speech compared to written text. . %Spoken language is inherently noisy, characterized by disfluencies, incomplete or fragmented utterances, diverse accents, and the absence of punctuation and casing. These features significantly complicate the task of identifying and correcting grammatical errors in speech compared to written text. 
%In several initial studies, LSTM-based models trained on a large amount of written second language (L2) learner data were investigated for GEC, GED, and disfluency detection \cite{knill2019sged, lu2019impact, lu2019disfluency}. 
Traditionally, spoken GEC systems have adopted a cascaded pipeline architecture, typically consisting of an automatic speech recognition (ASR) module to transcribe audio into text, followed by a disfluency detection module to produce fluent transcriptions, and finally a GEC module to correct grammatical errors~\cite{lu2020spoken, lu-etal-2022-assessing}. While this approach has shown some effectiveness, it is often hindered by error propagation across stages, which can degrade overall system performance.

This was followed by end-to-end approaches powered by large speech foundation models %such as Whisper ~\cite{radford2023robust}, 
which promise to decrease the number of compounded errors~\cite{banno2024towards}. To address the problem of the scale of data needed to build such systems, \citet{qian2025scaling} explored data augmentation, and \citet{qian2025endtoendspokengrammaticalerror} describe a novel reference alignment process to reduce transcription errors which outperforms a fine-tuned multimodal LLM \citet{lu2025advancingautomatedspeakingassessment} for this task. %While their approach outperforms a cascaded baseline, it still under performs compared to using a fine-tuned Whisper model~\cite{qian2025endtoendspokengrammaticalerror}.

%To address these limitations, recent research has begun exploring end-to-end approaches powered by large speech foundation models such as Whisper~\cite{radford2023robust}. These models offer the potential to simplify the architecture and reduce compounded errors~\cite{banno2024towards}. However, training such models effectively remains challenging due to the scarcity of large-scale, high-quality annotated spoken language datasets. To overcome this bottleneck, \citet{qian2025scaling} recently proposed a pseudo-labeling strategy to automatically expand SGEC training data, resulting in improved performance for end-to-end systems. To further improve the precision of grammatical error feedback, \citet{qian2025endtoendspokengrammaticalerror} introduced a novel reference alignment process designed to eliminate hypothesized edits arising from fluent transcription errors. Additionally, the authors combined this method with the earlier pseudo-labeling approach and incorporated an edit confidence estimation mechanism to filter out low-confidence edits. These enhancements led to notable improvements.

Despite these advances, generating accurate and meaningful feedback from spoken input continues to be a significant challenge. The recent release of the Speak \& Improve Corpus~\cite{knill2024speak}, the first publicly available speech dataset annotated for grammatical errors, and its associated challenge~\cite{qian2024speak} represents a major milestone and is expected to catalyze further progress and innovation in the field. 

\subsection{Reading} 
Assistive technologies for reading in NLP primarily focus on {\bf Automatic Readability Assessment} and {\bf Automatic Text Simplification}. Readability Assessment refers to the task of assigning a reading level to a given text based on its language difficulty, to various target readers. Interest in this topic is almost a century old among the education researchers \cite[e.g.,][]{vogel1928objective} while the NLP research has an over two decade history  \cite{kevyn2014computational,vajjala2022trends}, and different approaches from feature based machine learning to deep learning methods have been studied. 
Recent adaptation of LLMs to this problem so far seems to indicate that task-specific fine-tuned models achieve better results than zero- or few-shot prompting of LLMs
\cite{naous-etal-2024-readme,wang-etal-2024-fpt,smadu-etal-2024-investigating}. However, other work demonstrates better agreement between LLM-generated reading level judgments and human evaluations \cite{trott-riviere-2024-measuring}, and \citet{rooein-etal-2024-beyond} argue for new prompt-based evaluation metrics switching from the traditional static evaluation metrics while using LLMs. 

Text Simplification refers to the task of generating text in a simpler, easier to understand language, given a more complex text (typically sentence-to-sentence). It is a well-studied area of research in NLP \cite[\textit{inter alia}]{alva2020data,vstajner2021automatic,chi2023,huang-kochmar-2024-referee} and the advent of LLMs resulted in a natural extension of this research.   %with research focusing on all aspects of system building including datasets \cite{alva-manchego-etal-2020-asset,shardlow-etal-2024-bea}, sentence simplification models \cite{alva2020data,martin2022muss,chi2023}, evaluation methods \cite{vasquez-rodriguez-etal-2021-investigating,huang-kochmar-2024-referee}, user studies \cite{agrawal2024text} and broader impact discussions \cite{vstajner2021automatic}. 
While \citet{engelmann-etal-2024-arts} propose to use LLMs to create datasets for text simplification research, several groups showed the effectiveness of few-shot, in-context learning for generating diverse simplifications in multiple languages \cite{kew2023bless,nozza2023really,scalercio-etal-2024-enhancing}. Human user studies show better comprehension with LLM simplified text \cite{guidroz2025llm} but also substantial variation among human judgements \cite{trott-riviere-2024-measuring}. In terms of modeling, some recent approaches utilize LLMs and multi-agentic workflows to explore document-level simplification, showing promising early results \cite{mo-hu-2024-expertease,fang-etal-2025-collaborative,qiang2025redefiningsimplicitybenchmarkinglarge}. There is also a growing interest in personalizing text simplification through preference learning \cite{gao2025evaluatingeffectivenessdirectpreference}, generating texts at multiple levels of simplification \cite{,farajidizaji-etal-2024-possible,barayan-etal-2025-analysing}, domain specific simplification \cite{zecevic-etal-2024-simplification}, and elaborative simplification \cite{hewett-etal-2024-elaborative}. 

\subsection{Tutoring}
Within the domain of general knowledge acquisition and tutoring, one of the most effective NLP-enabled tools are {\bf Intelligent Tutoring Systems (ITS)}, in particular, dialogue-based ITSs. ITS are defined as computerized learning environments that incorporate computational models and provide feedback based on students’ learning progress \cite{graesser2001intelligent}; for dialogue-based systems, such feedback and communication with the student are empowered by NLP models. Lack of individualized tutoring has been linked to less effective learning and increased learner dissatisfaction~\cite{brinton2014individualization, eom2006determinants, hone2016exploring}, particularly in large classroom settings. This has led to the development of pre-LLM ITSs~\cite{paladines2020}, including systems focused on misconception identification~\cite{graesser1999autotutor, rus2013recent}, model-tracing tutors~\cite{rickel2002collaborative, heffernan2008expanding}, constraint-based models~\cite{mitrovic2005effect}, and Bayesian network models~\cite{pon2004evaluating} across educational levels. LLM-powered ITS systems promise to offer more personalized, one-on-one tutoring, enabling equitable and pedagogically sound learning experiences, which have long been known to lead to substantial learning gains~\cite{bloom19842}. Methods such as prompting \cite{wang-etal-2024-bridging}, fine-tuning \cite{jurenka2024responsibledevelopmentgenerativeai}, and Reinforcement Learning from Human Feedback (RLHF)~\cite{team2024learnlm} have been used in state-of-the-art LLM-based ITSs, as they help to overcome the limitations of traditional systems by enabling more adaptive, generalizable, and effective tutoring models.  

One of the key limitations for ITSs is the scope and size of current educational datasets \cite{macina2023mathdial, wang-etal-2024-bridging, stasaski2020cima}. Thus, building large-scale, publicly available educational datasets for LLM pre-training and fine-tuning should be prioritized in the near future. The focus on domain-specific models optimized for educational tasks and methods and the development of methods to assess the long-term impact of LLM-driven tutoring on learners and educators, including analysis of pedagogical effectiveness and bias, should also be considered more closely. % and ethical concerns in dialog-based interactions are all open challenges. 

\section{Assessment Technologies} 
\label{sec:assessment}
The assessment of writing, speaking, reading, and tutoring relies on a set of overlapping principles and techniques. Although each modality has its own unique features, they are deeply intertwined, particularly in the use of textual analysis and information extraction methods. Many of these techniques, initially developed within the domain of writing assessment, have been adapted for use in both speaking, reading and tutoring contexts, which we summarize in this section.  %such as through the analysis of automatic speech recognition (ASR) transcriptions, comprehension responses, or pedagogical responses. 

\subsection{Writing}
The origins of \textbf{automated writing assessment (AWA)} date back to the 1960s with the introduction of Project Essay Grade,  \cite{page1966imminence, page1968}. %, an early system that assessed writing ability using only proxy features. Over the following decades, advances in NLP technologies have significantly enhanced the capabilities of AWA. 
with notable progress in the 1990s and early 2000s that saw the emergence of the  commercial systems such as e-rater\textregistered~\cite{burstein2002erater}, IntelliMetric{\texttrademark}~\cite{rudner2006evaluation}, and the Intelligent Essay Assessor{\texttrademark}~\cite{landauer2002intell}. In later years, Deep Neural Network (DNN) approaches have led to substantial progress ~\cite{alikaniotis2016automatic}.
%with a particularly pivotal moment marked by the introduction of transformer-based architectures~\cite{vaswani2017attention}. Notably, BERT~\cite{devlin2019bert} stands out as a watershed model -- often seen as delineating the boundary between the pre-LLM and post-LLM eras -- due to its profound influence on NLP and, by extension, AWA.
In particular, transformer-based models have achieved performance levels that surpass even human inter-annotator agreement~\cite{rodriguez2019language}. Comprehensive overviews on AWA can be found in \citet{klebanov2022automated} and \citet{li-ng-2024-automated, li2024automatedessay}. 
%Building on this momentum, the advent of LLMs such as OpenAI’s GPT series~\cite{brown2020language, gpt4} has introduced a new paradigm in both natural language understanding and generation.
%These models, with their unprecedented scale and emergent capabilities, have the potential to revolutionize AWA -- not merely in terms of scoring accuracy, but also in how assessments are delivered, interpreted, and experienced by users. Moreover, their intuitive interfaces and widespread accessibility are reshaping how students, educators, and institutions interact with automated assessment tools.

Recent studies looked into the usefulness of LLMs for the assessment of second language (L2) writing, obtaining promising results \cite{mizumoto2023gpt,yancey2023rating}. 
%More recently, \citet{banno2024gpt} and \citet{stahl-etal-2024-exploring} investigated LLMs for analytic assessment and feedback, showing that they also have a good degree of understanding of individual aspects of language proficiency.
In line with \citet{liusie-etal-2024-llm}’s observation that LLMs tend to perform better at comparative rather than absolute assessment, \citet{cai2025rankthenscoreenhancinglargelanguage} proposed a combined ranking-and-scoring framework that outperforms standard prompt-based approaches. While most of the writing assessment research focused on evaluating the language proficiency aspect, a substantial amount of NLP research also focused on content assessment, in the form of \textbf{short answer scoring} \cite{burrows2015eras}. LLM based research on this topic is still emerging and recent studies so far conclude that zero/few-shot learning with LLMs fares poorly compared to traditional fine-tuning approaches for this task \cite{chamieh-etal-2024-llms, melloetallak25}. 

\subsection{Speaking}
Research on \textbf{automated speaking assessment (ASA)} began with relatively simple tasks, such as evaluating learners' ability to read individual words or sentences ~\cite{bernstein1990automatic, cucchiarini1997automatic, franco2000sri}. A significant milestone in this field was the development of SpeechRater system, which broadened the scope of automated assessment to include both spontaneous and read speech \cite{xi2008speechrater}. Recent years have seen significant advancements in the field through the adoption of DNN approaches~\cite{qian2012use}, and end-to-end neural-based methods have outperformed traditional systems such as SpeechRater~\cite{chen2018end}. A comprehensive survey of ASA can be found in \citet{zechner2019automated}. %, leveraging a wide range of features related to pronunciation, fluency, vocabulary, and grammar \cite{xi2008speechrater, zechner2009automatic, higgins2011three}. 

Pre-trained language models have contributed to further progress in ASA~\cite{raina2020universal, wang2021} and recent research has explored speech embedding representations for applications such as mispronunciation detection and diagnosis~\cite{wu21, xu2021mispron}, automatic pronunciation assessment~\cite{kim2022pron}, and the evaluation of proficiency across both monologic~\cite{banno23_slate, park2023multitask} and conversational~\cite{mcknight23_slate} data.

The application of speech-based LLMs in this domain is still in its early stages. \citet{fu2024pronunciationassessmentmultimodallarge} developed a speech LLM for L2 assessment that achieved competitive performance, albeit limited to the specific task of pronunciation scoring. With respect to holistic assessment, \citet{ma2025assessmentl2oralproficiency} recently explored the application of Qwen2-Audio~\cite{chu2024qwenaudio2}, for ASA in both zero-shot and fine-tuned settings. In a recent related study, \citet{banno2025naturallanguagebasedassessmentl2} demonstrated that integrating analytic proficiency descriptors with a zero-shot, text-based LLM applied to automatic transcriptions outperforms a BERT-based grader fine-tuned for the task, and achieves competitive performance compared to fine-tuned speech-based LLMs. This appears to be a promising direction to pursue in future research on ASA. 

\subsection{Reading}
One commonly studied problem in NLP in the area of reading assessment is the generation of reading comprehension questions. Question generation research in educational NLP and AI community in general addressed different scenarios from form-focused questions (e.g., to check grammatical knowledge) to more content-focused reading comprehension questions, using a range of methods from syntactic structures to neural language models \cite{kurdi2020systematic,perkoff-etal-2023-comparing,uto-etal-2023-difficulty,al2024review}. LLMs were used for question generation in math domain \cite{christ-etal-2024-mathwell,scarlatos-etal-2024-improving} and for personalized question generation in general \cite{xiao-etal-2023-evaluating,sauberli-clematide-2024-automatic}. Although English is the dominant language for research on this topic, cross-lingual transfer approaches have also been explored, and \citet{hwang-etal-2024-cross} show that smaller fine-tuned language models can achieve comparable performance to larger language models on this task. 

While the past research was restricted to a smaller set of datasets, the advent of LLMs resulted in approaches to benchmark construction and generation of questions at various difficulty levels according to a pre-existing taxonomy \cite{chen-etal-2024-dr,scaria-etal-2024-good}, and towards the development of novel evaluation approaches for automatically generated questions~\cite{moon-etal-2024-generative,pmlr-v264-deroy25a}. \citet{flor2025} presents an elaborate summary of automatic question generation research from traditional rule based methods to generative AI in a series of articles, which can serve as a good reference for those interested in further study. Item Response Theory (IRT, \citet{irt4nlp2024}), which has been widely used in psychometrics, has been explored in the context of vocabulary assessment \cite{ehara2012mining,ehara2018building} and developing questions for general language assessment \cite{settles-etal-2020-machine} in the past in NLP research. However, research combining automatic question generation with IRT is yet to emerge in NLP research, and appears to be a promising area where LLMs can help.    

\subsection{Tutoring}
Tutoring systems have long served as embedded assessment technologies, using learner interactions to evaluate understanding and guide instruction. Early systems like PLATO used rule-based feedback and simple branching logic for assessment and remediation \cite{woolf2010building}. Later, ITSs %such as SOPHIE and GUIDON 
incorporated expert system models and student diagnostic models to reason about domain knowledge and identify misconceptions \cite{listsknowledge}. By the 1990s, cognitive tutors like the Algebra Tutor employed cognitive models combined with model tracing and knowledge tracing to perform fine-grained, real-time skill assessment \cite{anderson1995cognitive}. Other ITS approaches such as AutoTutor utilize NLP models and dialogue-based reasoning to assess deeper conceptual understanding~\cite{nye_graesser_hu_2014}.

Recent advances in LLMs have transformed tutoring systems into flexible, multi-modal assessment environments. LLM-based platforms like Khanmigo \cite{shetye2024evaluation} and Google's LearnLM \cite{jurenka2024responsibledevelopmentgenerativeai, team2024learnlm} leverage generative AI to assess learner responses in natural language, interpret comprehension across reading, writing, and speaking tasks, and adapt instruction accordingly. Unlike traditional ITSs, LLMs enable open-ended, personalized feedback across diverse learning tasks, integrating instruction and assessment seamlessly \cite{venugopalan2025combining, wang2025llm}.

Despite their potential, LLM-based tutoring systems often lack rigorous validation linking their assessments to learning outcomes \cite{macina-etal-2023-opportunities}. Few studies have examined their diagnostic accuracy \cite{maurya-etal-2025-unifying}, adaptability across diverse learners \cite{wang2024tutor}, or long-term impact on knowledge retention \cite{kosmyna2025your}. Ethical concerns such as feedback bias and transparency also remain underexplored \cite{mvondo2023exploring}. Future research should develop standardized evaluation frameworks and investigate how LLM-driven assessments can be aligned with pedagogical goals.

%MathTutorBench: A Benchmark for Measuring Open-ended Pedagogical Capabilities of LLM Tutors
%https://arxiv.org/abs/2502.18940
%%Conversational ITS: BIPED: Pedagogically Informed Tutoring System for ESL Education 
%https://aclanthology.org/2024.acl-long.186/
Compared to assistive technologies, it appears that there are relatively fewer cases of LLMs' integration into assessment approaches, although it is clearly increasing. One reason could be that assessment is likely subject to more questions around reliability and validity of the models, considering the potential high stakes of the outcomes. Despite that, what we have seen so far shows how LLMs are increasingly being used in some of the common educational tasks traditionally studied in the NLP community. 

\section{New Directions Enabled by LLMs}
\label{sec:newdirections}
In this section, we turn to previously under-explored or new use cases enabled by LLMs across the four aspects (writing, speaking, reading, tutoring), for both assistive and assessment use cases. 

\paragraph{Content Generation:} A relatively new task, introduced with the advent of LLMs capable of fluent text generation in multiple languages, is educational content generation according to expert defined standards \cite{imperial-etal-2024-standardize}, for a specified grade level \cite{bezirhan2023automated,jin2025controllingdifficultygeneratedtext}, %describe using LLMs for automatic reading passage generation according to the given grade level specification, and \citet{logacheva2024evaluating} described the generation of contextual computer programming exercises with LLMs, to quote two examples covering different domains.
or for creation of evaluation and scaffolding exercises for different subjects \cite{xiao-etal-2023-evaluating,malik2025scaffolding}. One interesting question to extend this line of research further could be on-the-fly content generation given a topic, grade and standard specification, and target audience.

\paragraph{Multi-modal Interaction:} Text has been the dominant form of input in the development of educational technology applications. However, with multi-modal LLMs, some recent research explored other modes of interaction. Curating multi-modal content for education \cite{chaturvedi-2024-llms}, multi-modal question generation \cite{luo-etal-2024-chain}, end-to-end spoken language grammatical error correction \cite{banno2024towards}, low-resource language learning app development \cite{chu2025ataigi}, supporting listening assessment \cite{aryadoust2024investigating}, and evaluating handwritten exams \cite{liu2024aiassistedautomatedshortanswer} are some recent examples. Given these diverse use cases, and given that human learning can be considered multi-modal as we gain information from multiple forms of content, modeling of multi-modal interactions in human learning and multi-modal content generation can be considered challenging and useful future possibilities to study. 
%{\em TODO}: Add more about multi-modal systems for other domains, especially with ITS.

\paragraph{Synthetic Data Generation for Fine-tuning:} Synthetic data is increasingly being used at various stages of LLM training and fine-tuning pipelines, and education domain also started to see some new use cases for synthetic data such as aiding the development of educational chatbots and tutoring systems \cite{wang-etal-2024-book2dial,fateen2024developing}, spoken GEC~\cite{karanasou2025augmentation}, development of benchmark datasets for educational applications \cite{engelmann-etal-2024-arts,xu2025edubench}, and using LLMs as proxies for piloting educational assessments \cite{säuberli2025llmspsychometricallyplausibleresponses}. Considering the advantages synthetic data provides in terms of alleviating the need for labeled training data, exploring the limits and limitations of LLM-based synthetic data generation approaches for educational applications would be an important direction for the future.  

\paragraph{LLM Agents for Education:}  When LLMs are combined with components such as memory, tool use, and planning to solve complex tasks, they are referred to as \textit{LLM agents} \cite{chu2025llm, tran2025multi}. In an educational context, these additional components enable real-time adaptation, access to external resources, and planning of tailored learning paths, among other capabilities. At a high level, such agents function either as \textit{pedagogical agents} or \textit{domain-specific educational agents} \cite{chu2025llm}. Pedagogical agents imitate tutors to assist students or instructors in tutoring sessions and simulate students for tasks such as piloting exam questions or training tutors. Furthermore, multiple agents can operate simultaneously in multi-agent setups like in CAMEL \cite{li2023camel}, AutoGen \cite{wu2023autogen}, and PitchQuest \cite{mollick2024ai} to develop educational prototypes or solve complex problems. Domain-specific educational agents assist with learning in subjects such as science, languages, or professional development for specific domains. However, beyond general risks such as safety, hallucinations, and bias, the responses of the current state-of-the-art models are often not grade-appropriate \cite{Srivatsa-etal-Simulate}, may diverge from the learning path, conflate user roles, or enter conversational loops (amplified in multi-agent settings) \cite{li2023camel, chu2025llm, tran2025multi}. In summary, this research direction holds huge promise, but key limitations must be addressed when deploying these systems in sensitive domains like education.
% \paragraph{Agentic AI:} A lot of promise and novel directions allowed by implementing agents representing students on the one hand and tutors on the other. Overview relevant papers~\cite{chu2025llm,li2023camel,mollick2024ai,tran2025multi}.

\paragraph{Educational Human-LLM Collaborative Systems:} Human-LLM collaborative systems leverage the complementary strengths of humans and LLMs to improve performance in tasks such as data annotation, problem-solving, and decision-making across domains like education and healthcare \cite{yang-etal-2024-human, fragiadakis2024evaluating}. In education, LLM-powered systems have been employed to support both \textit{single-turn interactions} (e.g., answering questions, explaining steps) \cite{gao2024meta, hashir2024automatic} and \textit{multi-turn interactions} (e.g., Tutor-Copilot \cite{wang2024tutor}, GPTeach \cite{markel2023}). These systems can deepen learner understanding and assist novice tutors in improving their teaching skills and qualities. They are not free from challenges, though. Such systems often lack interpretability, making it hard to trust AI outputs \cite{yang-etal-2024-human}. They may prioritize correctness over pedagogical goals like conceptual understanding and learner support \cite{macina-etal-2023-opportunities}. LLMs also struggle with ambiguity, personalization, and maintaining context in extended interactions, and they rarely offer adaptive feedback tailored to learners’ evolving needs or emotions \cite{maurya-etal-2025-unifying}. Addressing these challenges is essential for building reliable and effective educational human-LLM collaborative systems.

%\paragraph{Human-computer interaction:} We are moving closely to the field of human-computer interaction (HCI), as the new challenges are around how to explore and benefit from collaboration of AI systems and real learners. There are examples to overview~\cite{markel2023,wang-etal-2024-bridging}.

\paragraph{Educational Value Alignment:} 
% Structure: (1) intro: success stories of LLM with alignment, (2) impact of alignment on education space, (3) what are grounding for the educational LLMs, (4) limitations & potential direction for the future research
% Current focus on ITS and needs to be updated for other applications
% Paragraph is inspired by Prof. Kochmar's initial draft
Alignment with human preferences is a key driver to the success of state-of-the-art LLMs \cite{ouyang2022training, yao2023instructions}. This ranges from the development of general values-based LLMs \cite{guo2025deepseek, team2024learnlm} to models tailored to specific age groups or domains \cite{nayeem2024kidlm, chen2023meditron}. These advancements have also significantly influenced the educational domain, leading to the development of education-specific LLMs such as LearnLM \cite{team2024learnlm, jurenka2024responsibledevelopmentgenerativeai} and pedagogical tutors \cite{dinucu2025problem}. These LLMs are grounded in pedagogical values \cite{team2024learnlm, maurya-etal-2025-unifying} and draw on decades of research in the learning sciences to generate pedagogically rich datasets, which are subsequently used for instruction tuning and fine-tuning. These specialized models have proven effective across a wide range of educational applications. However, an open research question remains -- ``\textit{What should we align with?}'' \cite{yao2023instructions}, which directly affects LLM performance. 

Specifically, in the case of tutor LLMs, there is currently no consensus among researchers regarding the key pedagogical principles and associated teacher moves that lead to effective learning \cite{team2024learnlm}. 
Future research should explore the core educational values that need to be integrated to enable the development of more effective educational models.

%\paragraph{Human Alignment:} Alignment of LLMs with human values (the general InstructGPT approach to be referenced here) led to the integration of similar ideas into educational applications. This is the most prominent in: (1) LLMs that are being aligned with the needs of specific types of learners -- for example, age-wise or subject-wise (an example to cite is KidLM); (2) integration of pedagogical principles and grounding in educational theories in ITS (examples include recent papers by ETH and our own unifying taxonomy). Specifically, alignment with human (or pedagogical) values is normally done via instruction tuning or fine-tuning, but the key question with which educational NLP is concerned is the definition of those principles and values that should be preserved in developed LLMs and educational applications (some examples need to be included here; for instance, a tutor / LLM should be "knowledgeable", but if integrated as a tutor in an educational dialogue, it should provide helpful feedback without revealing the answer). Other work to cite: Jurenka et al.  and LearnLM, with the pegagogical principles highlighted there.

%LLM-as-a-tutor in EFL Writing Education: Focusing on Evaluation of Student-LLM Interaction 
%https://aclanthology.org/2024.customnlp4u-1.21/

\section{Ongoing Challenges}% of LLMs and NLP in Education}
\label{sec:challenges}
So far, we have seen how LLMs enabled existing NLP research on educational applications, and also paved way for new use cases. With the growth in their usage in real-world educational scenarios and the potential for personalized education, a discussion about the challenges involved becomes inevitable. In this section, we discuss some of the the technical as well as broader application-related challenges in this area. 

\subsection{Datasets}
A lot of NLP research on educational applications relies on the existence of labeled datasets. For most of the tasks, such datasets are created by re-purposing existing online resources (e.g., using Wikipedia and Simple Wikipedia, websites such as Newsela for automatic text simplification), and this is not an exception compared to other areas of research in NLP. Carefully crafted datasets that are specifically developed for a particular task (e.g., grammatical error detection) are not rare, but hard to develop on a large scale. Datasets that consider target user input (e.g., those that contain learner feedback or outcome information) are even rare. Adding multilingual support to the mix makes dataset development across educational NLP tasks still more challenging. 

Although LLMs could offer better zero-shot, off-the-shelf performance for many tasks and languages today, and synthetic data generation with LLMs can address the data scarcity across languages to some extent, we would still need concerted efforts to build high quality educational datasets to develop and evaluate educational support LLMs across languages. Some recent research also reports poorer performance of LLMs across four education-related tasks beyond English and recommends verifying the LLM performance in the target language before deployment \cite{gupta2025multilingual}. \citet{imperial2025universalcefrenablingopenmultilingual}'s recent effort to consolidate multilingual language proficiency assessment datasets under one unified format and license is a welcome step in this direction. %We could expect to see more NLP research along these lines with the increasing use of LLMs in the development of educational applications that address various groups of target audience. 

%More focus on modular systems, which offer us more flexibility to adapt to target audiences quickly. 
%Explore multi-disciplinary collaboration to understand the requirements of different target user groups, and incorporate target domain expertise
%Develop datasets that suit these needs, new evaluation criteria (and standards to expand to other user groups), etc.

\subsection{Evaluation}
Across different NLP tasks involving the use of LLMs in the education domain, evaluation challenges have been widely discussed, along with a comparison between automated and human evaluation \cite{horbach-etal-2020-linguistic,vasquez-rodriguez-etal-2021-investigating,agrawal2024,kobayashi-etal-2024-large}.
While most of the discussions around evaluation focused on the task specific aspects, for technologies such as tutoring systems, a multi-dimensional view of evaluation is necessary. 

Traditional evaluation of teacher effectiveness has relied on artifacts, portfolios, self-reports, and student feedback \cite{goe2008approaches}. More recently, text generation metrics %NLG metrics like BLEU \cite{papineni2002bleu}, BERTScore \cite{zhang2019bertscore}, and DialogRPT \cite{gao2020dialogrpt} 
are being explored to assess ITS or AI tutor responses. However, while effective for measuring coherence and fluency, these domain-agnostic metrics often miss deeper pedagogical aspects, depend on gold references, and can be gamed by generic responses \cite{tack2022aiteachertestmeasuring}. Efforts to capture pedagogical effectiveness more directly have included human evaluations and tailored frameworks defining specific strategies, but these face challenges such as subjectivity, lack of standardization, and limited scope \cite{jurenka2024responsibledevelopmentgenerativeai}. Some potential directions for evaluating tutor responses were recently proposed by \citet{maurya-etal-2025-unifying}, where a unified taxonomy was introduced to measure the quality and appropriateness of these responses. However, these rely on human evaluation, which is non-scalable and is typically conducted at the utterance level rather than the conversation level. So far, using LLMs as proxy judges shows promise but still falls short in reliably evaluating complex pedagogical traits \cite{gu2024survey}.

Overall, NLP research is understandably model-focused and that impacts the way we evaluate. But, user focused evaluations are emerging. For example, some recent research points to the {\em mismatch between the user needs and model availability} in the context of graded content generation \cite{asthana-etal-2024-evaluating,kim-etal-2024-considering}. Future NLP research could consider a user-first  rather than a dataset- and model-first approach in developing standardized evaluation methodologies for LLM-based educational applications. 

\subsection{Ethical Issues}
 We found the discussion around the ethics of using LLMs in education emerging only recently in the NLP community \cite[e.g.,][]{hamalainen-2024-legal} but there has been some thought in this direction in the broader education technology and assessment community. \citet{yan2024practical} discuss the ethical implications of the increasing use of LLMs in education considering a range of use cases, and identify transparency for educational stakeholders (teachers, students, parents), privacy, support across languages, and fairness across population groups as the main ethical concerns surrounding the use of LLMs in education, calling for better reporting standards from empirical research that uses LLMs to develop new solutions. This issue of reporting standards is perhaps of most direct relevance to the NLP community. %, considering that our main area of focus is in developing new approaches, but not in studying end-user usage. 

From an assessment perspective, some recent work discussed the implications of the usage of LLMs in education to academic integrity \cite{de2024can,leppänen2025largelanguagemodelschanging} and fairness \cite{yamashita2025exploring} and the language assessment community calls for a collaboration between model developers, test creators and subject matter experts, psychometricians and the AI research community to develop education-specific standards for using AI in assessment to ensure reliability and fairness \cite{bolender2023criticality,voss2023use,xi2023advancing}. %Complementing these discussions, \citet{yamashita2025exploring} examined the fairness of GPT-based AWA across demographic groups. They found that while GPT-4o produced unbiased ratings with respect to gender and socioeconomic status, it exhibited notable bias by race/ethnicity, highlighting the need for greater scrutiny of LLM-based assessment tools.

\paragraph{} Hallucination is a well-known concern with LLMs, and educational use cases are not immune to that. Some recent research on text simplification \cite{hewett-etal-2024-elaborative,zecevic-etal-2024-simplification} pointed to how the tendency to hallucinate increases as the task gets more complex such as generating in a specific domain or in a new language, for example.
%Interpretability issues: it may be important for some of the usecases we discussed. 
While this discussion about the challenges is non-exhaustive, it broadly highlights some of the general task-agnostic issues related to the use of LLMs and NLP in education. 

\section{Conclusions}
\label{sec:summary}
We presented how LLMs are integrated into existing research on the NLP-driven educational applications, and how they opened up new directions of research. Our study shows that LLMs lead to several interesting new developments which hold a lot of promise for the future in terms of both effective performance as well as inclusive development of applications addressing different languages and population groups. However, there are also several ongoing challenges related to available data, evaluation, and ethical concerns. As suggested by others, we envision an increase in inter-disciplinary collaboration between NLP researchers, domain experts and educators in leading to the development of better assistive and assessment technologies to support students and teachers in the future. We hope this paper would serve as a good starting point for NLP researchers about the state of the art in the educational applications of NLP using LLMs and what lies ahead. 

\section*{Limitations}
\label{sec:limitations}
%Limitations of this paper
We perceive two primary limitations to this paper: (a) Since our goal in this paper is to provide an overview of what lies ahead, we did not provide an exhaustive survey of the current state of the art. We focused largely on post-LLM research in this area, pointing to relevant surveys for pre-LLM approaches; and (b) We have also primarily restricted ourselves to NLP publication venues, citing research from other related disciplines to a much smaller extent. Our observations and conclusions drawn in this paper should be considered along with these limitations. However, we provide an extensive, although by no means an exhaustive, list of additional readings grouped by the four dimensions -- writing, speaking, reading and tutoring -- in the Appendix Tables ~\ref{tab:refs_writing}--\ref{tab:refs_tutoring}, for those interested in exploring these topics further. 

\section*{Ethical Considerations}
The study does not involve the use of any datasets with ethical concerns or training of AI models with potential ethical issues. Hence, we do not anticipate any significant risks associated with this work.

%\section*{Acknowledgements}

\bibliography{tacl2021,anthology}
% \bibliographystyle{acl_natbib}

%\iftaclpubformat

%\onecolumn

\appendix

\section{Background} 
\label{sec:relatedwork}
Post ChatGPT, there has been a lot of interest in the role of LLMs, and more generally, generative AI, in education. Several surveys and opinion pieces were published in the past two years outlining the potential use cases, ethical implications, and challenges of using LLMs in education, primarily by research groups focusing on the use of technology in education \cite{dempere2023impact,fuchs2023exploring,kasneci2023chatgpt,yan2024practical}. However, none of them take a closer look at methods (LLM-based or not) used to address the different use cases of LLMs in education, which is of primary interest with NLP. 

NLP research has a long history of working on problems of relevance to the education domain, such as automated language assessment and grammatical error correction, among others. \citet{burstein2009opportunities} and \citet{meurers2012natural} give an overview of how NLP plays a role in the development of computational tools for language learning and assessment, and more recently, \citet{vajjala2024generative} has explored the role of generative AI in language learning technology. But all three are primarily directed towards non-NLP audience and we are not aware of any such discussion directed towards an NLP audience, focusing on the recent developments in generative AI and large language models.   

\citet{caines2023application} discuss the role of LLMs specifically in language teaching and assessment technologies from an NLP standpoint, with a particular focus on writing assessment, while not addressing speaking assessment. However, the use of language in the educational application space is not limited to language learning alone, and it encompasses other areas such as content learning, as well as assessment and learning support for other subjects. In this paper, we aim to give a broader perspective on the relevance of LLMs for the development of language-based educational applications in general, and how other NLP methods fit into the current scenario. Where relevant, we will also summarize the perspectives on the use of NLP and LLMs in this area from other related disciplines.

\section{Additional References}

%Perhaps add a paragraph on the methodology for choosing papers?: aclweb, 2023-25 to identify new use cases.

%detailed tables about papers considered while writing Sections 3-4. 

%\fi

We provide more detailed references for additional reading on specific topics in this section, grouping them along the four dimensions: writing, speaking, reading, and tutoring, in the Tables ~\ref{tab:refs_writing}--\ref{tab:refs_tutoring}.

\begin{table*}
  \centering
  \begin{tabular}{|c|p{12cm}|}
    % \hline
% \textbf{Type of paper} & \textbf{References} \\ \hline
\hline 
\multicolumn{2}{c}{\textbf{Written GEC \& GED}} \\ \hline
Surveys & \newcite{bryant-etal-2023-grammatical} \\ \hline 
Datasets & \newcite{yannakoudakis-etal-2011-new,dahlmeier-etal-2013-building,dale-etal-2012-hoo,ng-etal-2013-conll,ng-etal-2014-conll,mohit-etal-2014-first,rozovskaya-etal-2015-second,napoles-etal-2017-jfleg,bryant-etal-2019-bea,rozovskaya-roth-2019-grammar,koyama-etal-2020-construction,naplava-etal-2022-czech,masciolini-etal-2025-multigec} \\ \hline 
Evaluation & \newcite{dahlmeier-ng-2012-better,felice-briscoe-2015-towards,napoles-etal-2015-ground,napoles-etal-2016-theres,bryant-etal-2017-automatic,choshen-etal-2020-classifying,belkebir-habash-2021-automatic} \\ \hline 
\makecell[t]{Pre-LLM\\Approaches} & \newcite{chodorow-etal-2007-detection,kochmar-etal-2012-hoo,felice-etal-2014-grammatical,junczys-dowmunt-grundkiewicz-2014-amu,junczys-dowmunt-grundkiewicz-2016-phrase,junczys-dowmunt-etal-2018-approaching,yuan-etal-2019-neural,malmi-etal-2019-encode,stahlberg-kumar-2020-seq2edits,kaneko-etal-2020-encoder,omelianchuk-etal-2020-gector,mallinson-etal-2020-felix,katsumata-komachi-2020-stronger,mallinson-etal-2022-edit5,alhafni-etal-2023-advancements,zhou-etal-2023-improving-seq2seq,mesham-etal-2023-extended,alhafni-habash-2025-enhancing} \\ \hline
\makecell[t]{LLM\\Approaches} & \newcite{fang2023chatgpt,coyne2023analyzingperformancegpt35gpt4,wu2023chatgptgrammarlyevaluatingchatgpt,loem-etal-2023-exploring,raheja-etal-2023-coedit,davis-etal-2024-prompting,katinskaia-yangarber-2024-gpt,omelianchuk-etal-2024-pillars,kaneko-okazaki-2023-reducing,katinskaia-yangarber-2024-gpt,raheja-etal-2024-medit,omelianchuk-etal-2024-pillars,kobayashi-etal-2024-large,mita-etal-2024-towards} \\ \hline
\multicolumn{2}{c}{\textbf{GEE}} \\ \hline
Datasets & \newcite{nagata-2019-toward,nagata-etal-2020-creating,pilan-etal-2020-dataset,lopez-cortez-etal-2024-gmeg,kobayashi-etal-2024-large} \\ \hline
\makecell[t]{Pre-LLM\\Approaches} & \newcite{nagata-2019-toward,pilan-etal-2020-dataset} \\ \hline 
\makecell[t]{LLM\\Approaches} &  \newcite{lopez-cortez-etal-2024-gmeg,kobayashi-etal-2024-large} \\\hline 
\multicolumn{2}{c}{\textbf{Automatic Writing Assessment}} \\ \hline
Surveys & \newcite{shermis2003automated, shermis2010automated, shermis2013handbook, ke2019automated, beigman-klebanov-madnani-2020-automated, klebanov2022automated, li2024automatedessay, li-ng-2024-automated, shermis2024routledge} \\ \hline
Datasets & \newcite{granger1993international,yannakoudakis-etal-2011-new, blanchard2013toefl11, geertzen2013automatic, ishikawa2013icnale, ostling-etal-2013-automated, boyd-etal-2014-merlin, rakhilina-etal-2016-building, horbach-etal-2017-fine, mathias-bhattacharyya-2018-asap, glaznieks2023kolipsi, marinho2021, habash-palfreyman-2022-zaebuc, naismith2022university, crossley2023english, nicholls2024write, imperial2025universalcefrenablingopenmultilingual} \\ \hline
Evaluation & \newcite{williamson2012framework, buzick2016comparing, rotou2020evaluations} \\ \hline
Pre-LLM Approaches & \newcite{burstein2002erater, landauer2002intell, rudner2006evaluation, yannakoudakis-etal-2011-new, chen-he-2013-automated, zesch-etal-2015-task, alikaniotis2016automatic, vajjala2018automated, rodriguez2019language, yang-etal-2020-enhancing, wang-etal-2022-use} \\ \hline
LLM Approaches & \newcite{mizumoto2023gpt, yancey-etal-2023-rating, banno-etal-2024-gpt, song2024automated, stahl-etal-2024-exploring, atkinson2025llm, cai2025rankthenscoreenhancinglargelanguage, yamashita2025exploring} \\ \hline

\multicolumn{2}{c}{\textbf{Short Answer Scoring}} \\ \hline
Surveys & \newcite{ziai2012short,burrows2015eras} \\ \hline
Datasets & \newcite{meurers-etal-2011-evaluating1,ouahrani2020ar}\\ \hline
Pre-LLM Approaches & \newcite{leacock2003c,uto2020automated,horbach2024crosslingual} \\ \hline
LLM Approaches & \newcite{chamieh-etal-2024-llms,melloetallak25} \\ 
\hline 
  \end{tabular}
  \caption{Additional references for writing tasks (assistive/assessment)}
  \label{tab:refs_writing}
\end{table*} %TODO: ADD FROM ASSESSMENT

\begin{table*}
  \centering
  \begin{tabular}{|c|p{10.5cm}|}
    % \hline
% \textbf{Type of paper} & \textbf{References} \\ \hline
\hline 
\multicolumn{2}{c}{\textbf{Readability Assessment}} \\ \hline
Surveys & \newcite{collins2014,vajjala2022trends} \\ \hline
Datasets & \newcite{paetzold-specia-2016-semeval, vajjala-lucic-2018-onestopenglish,shardlow-etal-2021-semeval,seiffe-etal-2022-subjective,naous-etal-2024-readme} \\ \hline
Evaluation & \newcite{vajjala-etal-2016-towards,todirascu-etal-2016-cohesive,vajjala-lucic-2019-understanding,shubi-etal-2024-fine} \\ \hline
Pre-LLM Approaches & \newcite{collins-thompson-callan-2004-language,pitler-nenkova-2008-revisiting,feng-etal-2010-comparison,vajjala-meurers-2012-improving,xia-etal-2016-text,nadeem-ostendorf-2018-estimating,azpiazu-pera-2019-multiattentive,deutsch-etal-2020-linguistic,lee-etal-2021-pushing,wilkens-etal-2024-exploring} \\ \hline 
LLM Approaches & \newcite{lee-lee-2023-prompt,nohejl-etal-2024-difficult,rooein-etal-2024-beyond,wang-etal-2024-fpt,smadu-etal-2024-investigating} \\ \hline
\multicolumn{2}{c}{\textbf{Text Simplification}} \\ \hline
Surveys & \newcite{siddharthan-2014,alva2020data} \\ \hline
Datasets & \newcite{zhu-etal-2010-monolingual,coster-kauchak-2011-simple,kauchak-2013-improving,hwang-etal-2015-aligning,xu-etal-2015-problems,xu-etal-2016-optimizing,kajiwara-komachi-2016-building,zhang-lapata-2017-sentence,sulem-etal-2018-simple,scarton-etal-2018-simpa,vajjala-lucic-2018-onestopenglish,saggion-etal-2022-findings,hayakawa-etal-2022-jades,ryan-etal-2023-revisiting,alhafni-etal-2024-samer,shardlow-etal-2024-bea,jiang-xu-2024-medreadme,saggion-etal-2024-lexical,qiu-etal-2024-complex,nagai-etal-2024-document} \\ \hline
Evaluation & \newcite{xu-etal-2016-optimizing,sulem-etal-2018-semantic,vasquez-rodriguez-etal-2021-investigating,alva2021,cardon-etal-2022-linguistic,huang-kochmar-2024-referee,agrawal2024} \\ \hline 
\makecell[t]{Pre-LLM\\Approaches} & \newcite{chandrasekar-etal-1996-motivations,elhadad-sutaria-2007-mining,zhu-etal-2010-monolingual,woodsend-lapata-2011-learning,wubben-etal-2012-sentence,kajiwara-etal-2013-selecting,shardlow-2014-open,xu-etal-2016-optimizing,paetzold-specia-2016-semeval,nisioi-etal-2017-exploring,zhang-lapata-2017-sentence,alva-manchego-etal-2017-learning,de-hertog-tack-2018-deep,stajner-nisioi-2018-detailed,guo-etal-2018-dynamic,maddela-xu-2018-word,zhao-etal-2018-integrating,gooding-kochmar-2018-camb,vu-etal-2018-sentence,gooding-kochmar-2019-recursive,surya-etal-2019-unsupervised,qiang:2020:lexical,martin-etal-2020-controllable,omelianchuk-etal-2021-text,maddela-etal-2021-controllable,qiang:2021:lsbert,hazim-etal-2022-arabic,martin-etal-2022-muss,sheang-etal-2022-controllable} \\ \hline
\makecell[t]{LLM\\Approaches} & \newcite{chi2023,nozza-attanasio-2023-really,kew-etal-2023-bless,trott-riviere-2024-measuring,scalercio-etal-2024-enhancing,mondal-etal-2024-dimsim,zecevic-etal-2024-simplification,tan-etal-2024-llm,hewett-etal-2024-elaborative,qiu-zhang-2024-label,zetsu-etal-2024-edit,asthana-etal-2024-evaluating,farajidizaji-etal-2024-possible,mo-hu-2024-expertease,barayan-etal-2025-analysing} \\ \hline 
\multicolumn{2}{c}{\textbf{Question Generation}} \\ \hline
Surveys & \newcite{kurdi2020systematic,al2024review,flor2025} \\ \hline
Datasets & \newcite{chen2018learningq}\\ \hline 
Evaluation & \newcite{horbach-etal-2020-linguistic,xiao2023evaluating,pmlr-v257-gorgun24a,pmlr-v264-deroy25a}\\ \hline
\makecell[t]{Pre-LLM\\Approaches}& \newcite[][ch4--ch9]{flor2025} \\ \hline
\makecell[t]{LLM\\Approaches} & \newcite[][ch10]{flor2025}, \newcite{al2024analysis,sauberli2024automatic,scaria2024good,kumar2024improving}\\ \hline 
  \end{tabular}
  \caption{Additional references for reading tasks (assistive/assessment)}
  \label{tab:refs_reading}
\end{table*}

\begin{table*}
  \centering
  \begin{tabular}{|c|p{10cm}|}
    % \hline
% \textbf{Type of paper} & References \\ \hline 
\hline 
\multicolumn{2}{c}{\textbf{Spoken GEC \& GED}} \\ \hline
Evaluation &  \newcite{lu-etal-2022-assessing, qian2025endtoendspokengrammaticalerror}\\\hline
Datasets & \newcite{izumi2004, caines-etal-2016-crowdsourcing, kim2024learnervoice, knill2024speak} \\ \hline
\makecell[t]{Pre-LLM\\Approaches} & \newcite{izumi-etal-2003-automatic, lee2011grammatical, knill2019sged, lu2019impact, lu2019disfluency, lu2020spoken,lu-etal-2022-assessing, banno2023b_slate_grammatical, banno2024towards, karanasou2025augmentation, qian2025scaling, qian2025endtoendspokengrammaticalerror} \\ \hline
\makecell[t]{LLM\\Approaches} & \newcite{lu2025advancingautomatedspeakingassessment} \\ \hline
\multicolumn{2}{c}{\textbf{Spoken Language Assessment}} \\ \hline
Surveys & \newcite{zechner2019automated} \\ \hline
Datasets & \newcite{menzel-etal-2000-isle,izumi2004, yoon2009construction,ishikawa2014design, baur2017overview, baur2018overview, zhao2018l2arctic,baur2019,ishikawa2019icnale,speechocean762,coulange-etal-2024-corpus,kim2024learnervoice,knill2024speak} \\ 
\hline 
\makecell[t]{Pre-LLM\\Approaches} & \newcite{bernstein1990automatic, cucchiarini1997automatic, townshend1998estimation, franco2000sri, xi2008speechrater, qian2012use, malinin2017, chen2018end, evanini2018improvements, craighead-etal-2020-investigating, raina2020universal, peng2021mispron, wu2021mispron, xu2021mispron, wang2021, kim2022pron, banno23_slate, mcknight23_slate, park2023multitask} \\\hline
\makecell[t]{LLM\\Approaches} & \newcite{fu2024pronunciationassessmentmultimodallarge, phan2024automated, banno2025naturallanguagebasedassessmentl2, ma2025assessmentl2oralproficiency, voskoboinik2025leveraging} \\ \hline
  \end{tabular}
  \caption{Additional references for speaking tasks (assistive/assessment)}
  \label{tab:refs_speaking}
\end{table*}

\begin{table*}
  \centering
  \begin{tabular}{|c|p{10cm}|}
    % \hline
% \textbf{Type of paper} & References \\ \hline
\multicolumn{2}{c}{\textbf{Intelligent Tutoring Systems}} \\ \hline
Surveys &  \newcite{paladines2020,wollny2021,wang2024largelanguagemodelseducation}
\\ \hline
Datasets & \newcite{stasaski-etal-2020-cima,caines-etal-2020-teacher,suresh-etal-2022-talkmoves,demszky-hill-2023-ncte,macina-etal-2023-mathdial} \\ \hline
Evaluation & \newcite{demszky-etal-2021-measuring,vasselli-etal-2023-naisteacher,jurenka2024responsibledevelopmentgenerativeai,maurya-etal-2025-unifying} \\ \hline
\makecell[t]{Pre-LLM\\Approaches} & \newcite{evers:2000,freedman-2000-plan,suraweera2022,graesser2004autotutor,graesser2006autotutor,weerasinghe2006individualizing,dzikovska-etal-2010-beetle,Mello2012,Cristobal:2013,nye_graesser_hu_2014,serban2020,macina-etal-2023-opportunities} \\ \hline 
\makecell[t]{LLM\\Approaches} & \newcite{tack2022aiteachertestmeasuring,tack-etal-2023-bea,vasselli-etal-2023-naisteacher,wang-demszky-2023-chatgpt,sonkar-etal-2023-class,lee2023learningteachingassistantsprogram,markel2023,daheim-etal-2024-stepwise,chowdhury2024autotutormeetslargelanguage,wang-etal-2024-backtracing,wang-etal-2024-bridging,denny2024generativeaieducationgaied,wang-etal-2024-book2dial,nie2025gptsurpriseofferinglarge,Srivatsa-etal-Simulate} \\ \hline 
  \end{tabular}
  \caption{Additional references for tutoring tasks (assistive/assessment)}
  \label{tab:refs_tutoring}
\end{table*}

\end{document}